\documentclass{article}
\usepackage{iclr2027_conference,times}

\usepackage{amsmath,amsfonts,bm}

\def\eqref#1{equation~\ref{#1}}

\def\1{\bm{1}}

\DeclareMathAlphabet{\mathsfit}{\encodingdefault}{\sfdefault}{m}{sl}
\SetMathAlphabet{\mathsfit}{bold}{\encodingdefault}{\sfdefault}{bx}{n}

\usepackage{hyperref}
\usepackage{url}
\usepackage{booktabs}
\usepackage{amsmath}
\usepackage{amssymb}
\usepackage{graphicx}
\usepackage{caption}
\usepackage{multirow}
\usepackage{wrapfig}
\usepackage{xcolor}

\title{\textsc{Ladder}: Graph-Guided Diffusion Language Models for Efficient Multi-Hop Reasoning}

\author{Senlei Zhang$^{1}$\footnotemark[2],~~Linhao Luo$^{2}$,~~Qian-wen Zhang$^{3}$,\\\textbf{Siyu An}$^3$,~~\textbf{Junnan Dong$^{3}$\footnotemark[1]},~~\textbf{Shuhao Zhang$^{1}$\footnotemark[1],~~Xing Sun$^3$}\\
$^1$Huazhong University of Science and Technology, $^2$Monash University, $^3$Tencent Youtu Lab\\
~~\texttt{\{hansonjdong\}@tencent.com}
}

\iclrfinalcopy

\begin{document}

\maketitle
\lhead{Under review as a conference paper at ICLR 2027}
\renewcommand{\thefootnote}{\fnsymbol{footnote}}
\footnotetext[2]{Work done during internship at Tencent Youtu Lab.}
\footnotetext[1]{Corresponding authors.}

\begin{abstract}
Graph Retrieval-Augmented Generation (GraphRAG) has remarkably enhanced large language models on complex reasoning by leveraging structured entity topologies. However, existing frameworks heavily rely on standard autoregressive language models where the nature of inherent sequential generation severely hinders overall inference efficiency. Inspired by Diffusion Language Models (DLMs) that offer massive parallelism via continuous refine-in-parallel decoding, we aim to accelerate GraphRAG in the discrete space. However, it remains non-trivial for two challenges. First, partially denoised drafts are highly dynamic and uncertain, making dynamic graph grounding non-trivial. Second, raw denoising states are inherently noisy and unstable, making synchronous graph retrieval and multi-hop aggregation computationally prohibitive. To this end, we present \textsc{Ladder}, a novel framework that bridges diffusion language modeling with GraphRAG through graph-guided parallel decoding. Specifically, $(i)$ we propose an event-driven self-clocking retrieval, inspired by our key insight that $88\%$ of target entities emerge early in the partially denoised state, leading final commitment by an average of $5.7$--$9.6$ steps. This mechanism dynamically triggers graph retrieval only when the set of graph-linkable entities expands, yielding an asynchronous self-clocking policy that bypasses learned gates or heuristic thresholds. $(ii)$ An incomplete-query graph propagation module is designed to process the newly emerging entity queries using a specialized graph foundation model, continuously aggregating multi-hop evidence to sharpen parallel predictions and accelerate overall decoding convergence. Extensive experiments on three challenging multi-hop QA benchmarks show that \textsc{Ladder} raises average exact match from $39.6\%$ to $45.2\%$ while achieving a $4.1\times$ latency reduction.
\end{abstract}

\section{Introduction}
Graph Retrieval-Augmented Generation (GraphRAG) has remarkably enhanced large language models (LLMs) on complex multi-hop reasoning by leveraging structured entity topologies~\citep{dong2023hierarchy,luo2025gfmrag,hipporag2,edge2024local,guo2024lightrag,sarthi2024raptor}. In multi-hop question answering, dependencies are rarely visible in a single lexical query; intermediate factual entities must be established before downstream evidence can be identified. Existing GraphRAG frameworks typically interleave search with autoregressive reasoning~\citep{trivedi2023interleaving,dong2023active,yao2023react}. Under autoregressive decoding, however, these mechanisms inherit a fundamental bottleneck: the search query for a subsequent hop cannot be formulated until all preceding reasoning tokens are sequentially generated, and even confidence-triggered variants~\citep{jiang2023active} still resolve one token position at a time. Consequently, retrieval remains a synchronous operation strictly on the inference critical path, severely hindering overall generation efficiency. Diffusion Language Models (DLMs)~\citep{austin2021d3pm,lou2024discrete,sahoo2024mdlm,nie2025large} introduce a promising paradigm shift by refining all response positions in parallel. Instead of building a committed prefix token-by-token, a DLM's intermediate state acts as a continuous, noisy draft of the entire reasoning trajectory. This structural property suggests a novel role for structured retrieval~\cite{an2026toward} that dynamic entity-document graphs can guide the DLM throughout its parallel denoising process, connecting emerging entities in the draft to multi-hop evidence and resolving prediction uncertainty on the fly. As evidence sharpens dependent position predictions, more positions clear the sampling threshold in parallel, enabling the trajectory to converge in significantly fewer steps.

However, integrating DLMs for efficient reasoning remains non-trivial due to two fundamental challenges. First, partially denoised drafts are highly dynamic and fluctuate throughout decoding. Retrieving statically from the initial question misses emergent evidence, whereas querying at every denoising step incurs severe latency and overwrites useful conditioning with near-duplicate graph states. Second, raw denoising states are inherently noisy and unstable. Attempting synchronous graph retrieval and multi-hop aggregation directly over raw, unverified token drafts introduces prohibitive computational overheads and risks cascading retrieval errors.

To this end, we present \textsc{Ladder}, a novel framework that bridges diffusion language modeling with GraphRAG through graph-guided parallel decoding. Our framework resolves the above tensions through two key innovations. Specifically, $(i)$ Event-Driven Self-Clocking Retrieval: inspired by our key insight that $88\%$ of target entities emerge early in the partially denoised state, leading final commitment by an average of $5.7$--$9.6$ steps, we propose an event-driven retrieval mechanism. Rather than relying on learned gates or confidence thresholds, \textsc{Ladder} dynamically triggers graph retrieval only when the set of graph-linkable entities expands. This yields an asynchronous self-clocking policy where each emerging entity serves both as a refresh event and as a concrete seed node for graph propagation. $(ii)$ Incomplete-Query Graph Propagation: to process the revisable, incomplete event queries generated during intermediate denoising, an incomplete-query graph propagation module is designed using a specialized graph foundation model~\citep{galkin2024ultra,luo2025gfmrag,zhu2021nbfnet} (GFM), building on GNN-based retrievers designed to tolerate incomplete or noisy graphs~\citep{mavromatis2025gnnrag}. We fine-tune the GFM with complementary discovery and retention objectives: one retrieves gold evidence still absent from memory, while the other preserves evidence retrieved at earlier events. An anchor term further ensures that this adaptation transfers zero-shot across unseen evaluation corpora.

Our main technical contributions are summarized as follows:
\begin{itemize}
\item \textbf{Graph-Guided Diffusion Decoding Paradigm.} We identify the sequential bottleneck of traditional GraphRAG and formulate graph retrieval as a parallel guidance mechanism for DLM denoising trajectories, accelerating decoding while improving answer accuracy.
\item \textbf{Event-Driven Retrieval Policy.} We establish entity lead time as a strong empirical foundation and introduce a gate-free, self-clocking policy that uses emerging entities as both asynchronous refresh triggers and graph propagation start nodes.
\item \textbf{Incomplete-Query Graph Foundation Model.} We design an event-conditioned graph propagation model trained with discovery and retention objectives to resolve revisable event queries, achieving robust zero-shot cross-corpus transferability.
\item Extensive experiments on three challenging multi-hop QA benchmarks show that \textsc{Ladder} raises average exact match from $37.9\%$ under retrieve-once to $45.2\%$, while reducing latency by $36\%$ relative to per-step retrieval.\end{itemize}

\section{Related Work}

\paragraph{Graph-enhanced retrieval augmented generation (GraphRAG).}
RAG~\citep{lewis2020rag} grounds a language model in external text, but dense
retrievers~\citep{karpukhin2020dpr,dong2024cost} score each document independently, so
evidence reachable only through a chain of intermediate facts is hard to surface
in one shot. Two lines address this multi-hop gap. Iterative retrieval lets the
model steer several rounds of search: IRCoT~\citep{trivedi2023interleaving,dong2024modality}
queries with each generated sentence, Self-Ask~\citep{press2023measuring} and
ReAct~\citep{yao2023react} decompose into sub-questions or tool calls, and
FLARE~\citep{jiang2023active} fires when token confidence drops. Yet every
round is a synchronous trip on the critical path, and under autoregressive
decoding the query for hop $k{+}1$ cannot form until hop $k$ is written, so
retrieval structurally lags generation. Graph-structured retrieval instead builds
explicit structure so that multi-hop evidence becomes a traversal:
HippoRAG~\citep{hipporag} runs personalised PageRank over an entity--document
graph, GraphRAG~\citep{edge2024local,dong2026deep} and LightRAG~\citep{guo2024lightrag} index
induced graph structure, and RAPTOR~\citep{sarthi2024raptor}, HippoRAG~2 and
Youtu-GraphRAG~\citep{hipporag2,dong2025youtu} refine it further. Because induced
graphs are noisy, learned propagation replaces symbolic traversal: GNN
retrievers~\citep{mavromatis2025gnnrag} tolerate missing edges, and graph
foundation models~\citep{galkin2024ultra} such as
GFM-RAG~\citep{luo2025gfmrag,zhu2021nbfnet} transfer to unseen corpora, which we
adopt as our retriever. All of these remain query-driven and schedule-agnostic:
they answer \emph{what} to retrieve given a query, and leave open \emph{when} the
query should be issued.

\paragraph{Diffusion language models.}
Discrete diffusion over token
sequences~\citep{austin2021d3pm,lou2024discrete} and its continuous-embedding and
masked variants~\citep{li2022diffusionlm,sahoo2024mdlm,nie2025large} decode by
refining an entire sequence, so an intermediate state is a noisy draft of the
whole response rather than a prefix. We exploit this global draft to couple
retrieval scheduling with graph structure: each newly emerging entity in the
draft determines both \emph{when} to refresh and \emph{which} seed node to
propagate from in the entity--document graph. This coupling exists only once
retrieval is graph-structured, and we further sharpen it by training the
retriever on the denoising events themselves.

\section{Observations: Entities Emerge Before They Are Committed}
\label{sec:motivation}

Our method is motivated by a measurable property of DLM decoding: graph-linkable
entities can appear in an intermediate draft before their token spans are
committed.

\paragraph{Measurement protocol.}
We collect denoising trajectories for $1{,}000$ test questions from each of
2WikiMultihopQA~\citep{ho2020constructing},
HotpotQA~\citep{yang2018hotpotqa}, and MuSiQue~\citep{trivedi2022musique}.
Dream-7B uses a confidence-threshold sampler: at each step, every masked
position whose predicted-token confidence exceeds a threshold $\tau_c$ is
committed and is never revised afterward. At step $t$, we extract entities
from the argmax proxy $r_t$: committed positions kept as-is, masked positions
filled by their current top-1 prediction (Eq.~\ref{eq:proxy}). Let
$\mathcal{C}_t$ be the positions committed under this rule by step $t$, and
let $\mathcal{J}(e)$ denote the final-response token positions of entity $e$.
We measure

\begin{equation}
t_{\mathrm{emerge}}(e)=\min\{t:e\in\Delta E_t\},\qquad
t_{\mathrm{commit}}(e)=\min\{t:\mathcal{J}(e)\subseteq\mathcal{C}_t\},
\end{equation}
\begin{equation}
\mathrm{LeadTime}(e)=t_{\mathrm{commit}}(e)-t_{\mathrm{emerge}}(e).
\end{equation}

This span-level definition requires all of an entity's token positions to be
committed, so it excludes transient proxy guesses without coupling an entity's
commitment to unrelated earlier positions.

\begin{figure}[t]
\centering
\includegraphics[width=\textwidth]{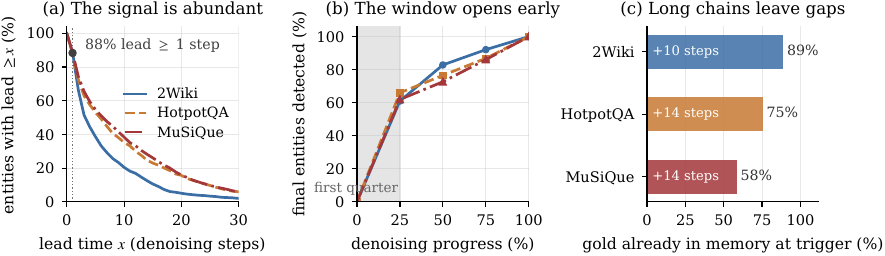}
\caption{Entities surface before their spans commit, early enough to be
useful. (a) Fraction of entities still uncommitted $x$ steps after surfacing
in $E_t$. (b) Fraction of final entities already detected after a given
trajectory fraction. (c) Fraction of triggers with all gold documents already
in memory; labels give the mean remaining steps. Section~\ref{sec:training}
targets the coverage loss visible here on long reasoning chains.}
\label{fig:lead}
\end{figure}

For a trigger at step $t$ of question $i$, Figure~\ref{fig:lead}c defines its
head start as $H_{i,t}=T_i-t$, where $T_i$ is the final denoising step. It
records whether $\mathcal{D}_i^\star\subseteq M_{t-1}$ at that trigger and
reports the mean of $H_{i,t}$ over trigger events.

\paragraph{Findings.}
Figure~\ref{fig:lead} and Appendix~\ref{app:lead} show that $88\%$ of final
entity spans are detected before commitment, with mean lead times of
$5.7$--$9.6$ denoising steps. More than $60\%$ of final entities appear in the
first quarter of the trajectory, leaving a useful window for evidence retrieval.
The window is not uniformly easy to exploit: the fraction of trigger events for
which all gold documents are already available falls from $88.9\%$ on 2Wiki to
$58.4\%$ on MuSiQue. This motivates an event-conditioned retriever that both
discovers missing evidence and preserves evidence found at earlier events.
Appendix~\ref{app:case} illustrates this behavior on one trajectory.

\section{Method}
\label{sec:method}

\begin{figure*}[t]
\centering
\includegraphics[width=\textwidth]{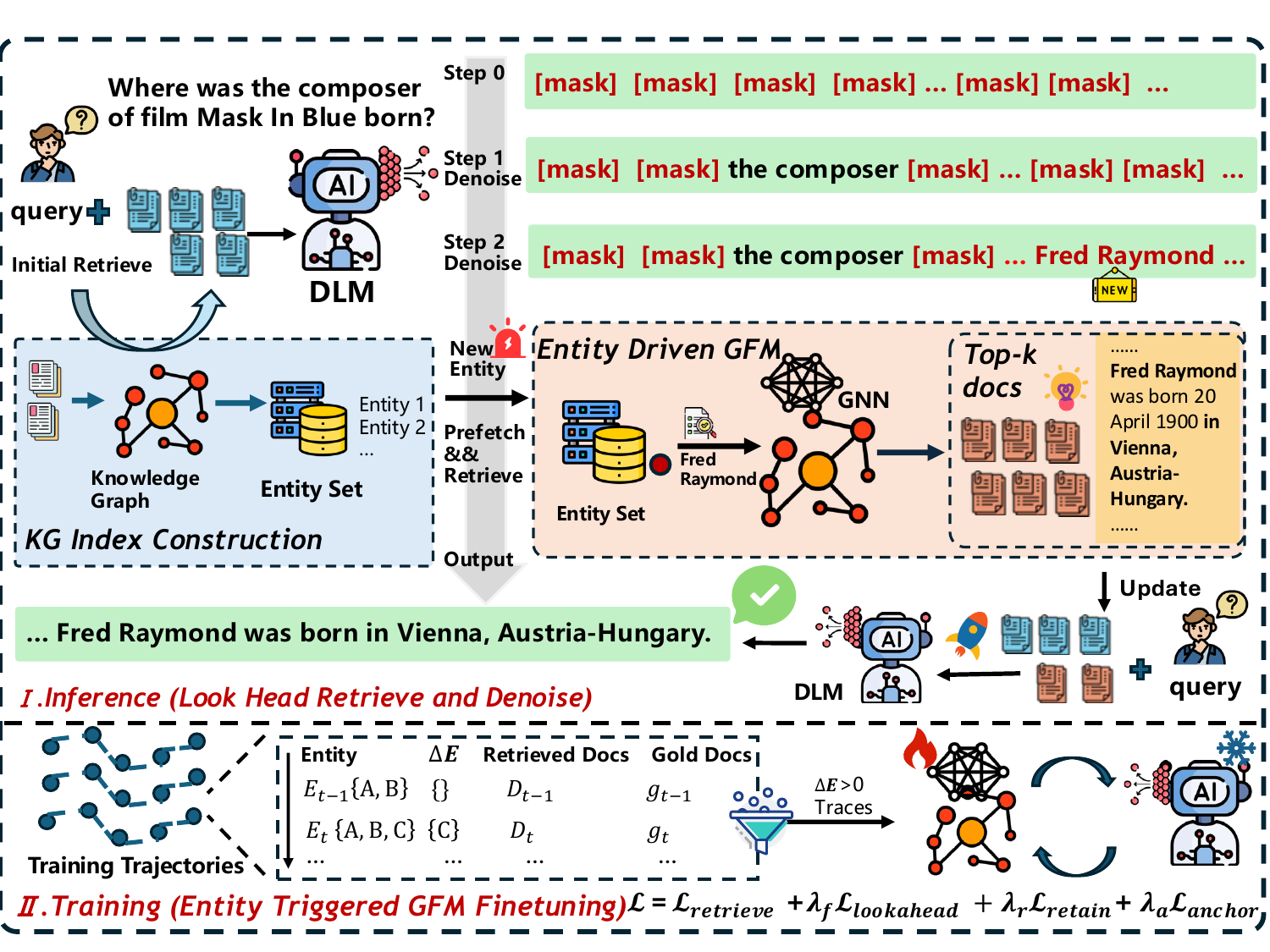}
\caption{Overview of \textsc{Ladder}. The DLM exposes a partially denoised response state at each step. Newly matched graph entities trigger GFM propagation and retrieve top-$K$ evidence, which conditions subsequent denoising updates. Equation~\ref{eq:total-loss} defines the complete retriever-training objective.}
\label{fig:method}
\end{figure*}

\textsc{Ladder} maintains a diffusion response state, a monotone set of
confirmed graph entities, and the evidence retrieved at the most recent event.
It uses the response state to expose a retrieval proxy, uses entity emergence to
schedule graph propagation, and trains the retriever on the resulting event
states.

\subsection{Denoising-time retrieval formulation}

Let $q$ denote a question, $\mathcal{G}=(\mathcal{V},\mathcal{R})$ an
entity--document graph, and $x_t$ the discrete diffusion state after denoising
step $t\in\{1,\ldots,T\}$. Unlike an autoregressive prefix, $x_t$ contains a
partially specified draft of the entire response. We therefore expose a proxy
response $r_t$ to the retriever before all of its tokens are committed. For a
masked position $j$, let $p_\theta(\cdot\mid x_t)_j$ be the DLM distribution and
$\hat{x}_{t,j}=\arg\max_v p_\theta(v\mid x_t)_j$. Inspired by the idea of using
an intermediate diffusion draft as a retrieval signal~\citep{juenger2026sardi},
we form the proxy by retaining committed tokens and, at every masked position,
filling in the current argmax guess $\hat{x}_{t,j}$, regardless of its
confidence:
\begin{equation}
 r_{t,j} =
 \begin{cases}
 x_{t,j}, & \text{position } j \text{ already committed},\\
 \hat{x}_{t,j}, & \text{position } j \text{ still masked}.
 \end{cases}
\label{eq:proxy}
\end{equation}
No threshold governs what the proxy exposes: the sampler's confidence
threshold decides only which tokens are committed to the final response, and has
no bearing on retrieval. The proxy is passed to the entity matcher of
Section~\ref{sec:trigger} in full, so retrieval is driven solely by which
entities it newly contains, the source of the lookahead signal measured in
Section~\ref{sec:motivation}.

At initialization we retrieve from the question alone. At subsequent steps, the
retrieval query is
\begin{equation}
 u_0=q, \qquad u_t=[q\mathbin{\Vert}r_t],\quad t>0,
\label{eq:query}
\end{equation}
where $[\cdot\Vert\cdot]$ denotes text concatenation. Given retrieved evidence
$D_t$, we rebuild the RAG prompt $P(q,D_t)$ before the next denoising update
(Appendix~\ref{app:prompt} gives the exact template and output format).
At a refresh, we retain the response token identities and commitment mask, but
re-tokenize the complete prompt with the updated passages. Dream-7B recomputes
RoPE positions over the refreshed full sequence; we use neither fixed context
slots nor cross-refresh KV caching. The next denoising update therefore
reconditions the retained response state on the new evidence.

\subsection{Entity-emergence scheduling}
\label{sec:trigger}

Running graph propagation for every syntactically different proxy is wasteful:
most changes in a diffusion state do not change its graph start nodes. We cast
scheduling as a discrete graph-state transition. At step $0$, we apply NER and
embedding-based entity linking to $q$ to obtain initial graph seeds $E_0$.
Thereafter, we match $u_t$ against the closed vocabulary of entity-node names in
$\mathcal{G}$. The matcher lowercases and tokenizes the text, then uses a greedy
longest-match $n$-gram lookup (up to six tokens by default); it therefore
requires neither an additional language model nor online entity linking. Let
\begin{equation}
 \widehat{E}_t=\operatorname{Match}(u_t;\mathcal{V}_{\mathcal{G}}),\qquad
 \Delta E_t=\widehat{E}_t\setminus E_{t-1},\qquad
 E_t=E_{t-1}\cup\widehat{E}_t .
\label{eq:entity-event}
\end{equation}
The initial linked seeds are included in $E_0$, and later exact matches are
canonical graph node identities. Retrieval is scheduled by
\begin{equation}
 a_t=\mathbb{I}\left[|\Delta E_t|\geq 1\right], \qquad t>0,
\label{eq:trigger-rule}
\end{equation}
with an unconditional retrieval at $t=0$. When $a_t=0$, we reuse the most
recent score vector and evidence set rather than recomputing query features,
graph propagation, or document ranking. The policy is self-clocking: the same
new entities determine both \emph{when} to refresh and \emph{where} propagation
starts. It has no learned gate, confidence calibration, or reward signal.

The entity state is monotone. A speculative mention may disappear as $r_t$ is
revised and later recur; removing it from $E_t$ would trigger redundant refreshes.
Accumulation makes each canonical entity trigger at most once and preserves it as
a graph seed despite surface-form fluctuations.

\subsection{Event-conditioned graph retrieval}
\label{sec:graph-retrieval}

We instantiate the retriever with GFM-RAG~\citep{luo2025gfmrag}. At an event,
the query encoder produces $h_t=h(u_t)$ and the confirmed entities define a
multi-hot graph-start mask $m(E_t)$. An NBFNet-style GNN yields raw entity scores
$z_t(e)$. With $A_{ed}\in\{0,1\}$ denoting the fixed entity--document incidence
matrix and $f_e=\sum_d A_{ed}$, the document score is
\begin{equation}
 z_t=f_\theta\bigl(\mathcal{G},h_t,m(E_t)\bigr),\qquad
 s_t(d)=\sum_e \frac{A_{ed}}{f_e}z_t(e),\qquad
 D_t=\operatorname{TopK}_{d\in\mathcal{D}}\;s_t(d).
\label{eq:gfm}
\end{equation}
Here $\mathcal{D}\subset\mathcal{V}$ is the document-node set and entities with
$f_e=0$ have zero weight. During fine-tuning, this dense aggregation scores all
documents and is differentiable; inference restores GFM-RAG's pretrained hard
Top-20 entity ranker before selecting the final top-$K$ documents. Consequently,
retrieval can reach multi-hop evidence with little lexical overlap with $u_t$.
Our default uses cumulative seeding with all entities in $E_t$; Appendix
\ref{app:ablations} compares it with new-seed propagation from $\Delta E_t$.
Between events, we reuse the most recent graph scores and evidence set.

The top-$K$ documents at event $t$ form the prompt evidence $D_t$ ($K=7$ in all
main experiments; Appendix~\ref{app:k-sensitivity} sweeps $K$). Separately, we define the historical retrieval set
$M_t=M_{t-1}\cup D_t$ for training targets and coverage analysis; it need not be
identical to the documents in the current prompt.

\subsection{Event-conditioned retriever training}
\label{sec:training}

A retriever pretrained on complete questions is mismatched to the event states
in Equation~\ref{eq:query}: these queries contain an incomplete, revisable
reasoning draft and a newly changed graph seed. We therefore collect
\emph{entity-event traces} only from training questions. Each row records
$(q,u_t,E_{t-1},\Delta E_t,E_t,M_{t-1},D_t,\mathcal{D}_i^\star)$ for a step with
$|\Delta E_t|>0$, where $\mathcal{D}_i^\star$ is the question's gold supporting
document set. Query embeddings are cached when traces are collected; the DLM,
the graph structure, and the text encoder remain frozen during retriever
fine-tuning.

For an event $t$, define the evidence that has not yet entered memory and the
gold evidence already present in memory as
\begin{equation}
 \mathcal{F}_t=\mathcal{D}_i^\star\setminus M_{t-1},\qquad
 \mathcal{R}_t=\mathcal{D}_i^\star\cap M_{t-1}.
\label{eq:targets}
\end{equation}
The negative pool is the deduplicated non-gold subset of the current hard
retrieval candidates and pre-event memory,
\begin{equation}
 \mathcal{N}_t=\bigl(D_t\cup M_{t-1}\bigr)\setminus\mathcal{D}_i^\star.
\label{eq:negatives}
\end{equation}
Thus, the current candidate contribution contains at most $K=7$ documents; we
use neither in-batch nor additional background negatives. For any positive set
$A$, we use the marginal listwise objective
\begin{equation}
 \ell_{\mathrm{list}}(A,\mathcal{N}_t)=-\log
 \frac{\sum_{d\in A}\exp(s_t(d)/\tau)}
 {\sum_{d\in A\cup\mathcal{N}_t}\exp(s_t(d)/\tau)}.
\label{eq:listwise}
\end{equation}
This set-level objective assigns probability mass to the annotated support set;
we do not use an average of per-positive losses. The current-evidence and
lookahead terms are
$\mathcal{L}_{\mathrm{retrieve}}=\ell_{\mathrm{list}}(\mathcal{D}_i^\star,\mathcal{N}_t)$
and $\mathcal{L}_{\mathrm{lookahead}}=\ell_{\mathrm{list}}(\mathcal{F}_t,\mathcal{N}_t)$.
The latter explicitly trains an early entity event to retrieve evidence not yet
seen by the agent. To avoid demoting correct earlier evidence when a new entity
appears, we use a hardest-negative hinge loss,
\begin{equation}
 \mathcal{L}_{\mathrm{retain}}=
 \frac{1}{|\mathcal{R}_t|}\sum_{d^+\in\mathcal{R}_t}
 \left[m-s_t(d^+)+\max_{d^-\in\mathcal{N}_t}s_t(d^-)\right]_+.
\label{eq:retain}
\end{equation}
A term is set to zero when its positive set is empty. Finally, with $\theta_0$
the pretrained GFM parameters, we optimize
\begin{equation}
 \mathcal{L}=\mathcal{L}_{\mathrm{retrieve}}
 +\lambda_f\mathcal{L}_{\mathrm{lookahead}}
 +\lambda_r\mathcal{L}_{\mathrm{retain}}
 +\lambda_a\mathcal{L}_{\mathrm{anchor}},
 \qquad
 \mathcal{L}_{\mathrm{anchor}}=\sum_\ell
 \frac{\|\theta_\ell-\theta_{0,\ell}\|_2^2}{|\theta_\ell|},
\label{eq:total-loss}
\end{equation}
where $|\theta_\ell|$ is the number of scalar parameters in tensor $\ell$.
We fine-tune all GFM GNN parameters using AdamW. Unless otherwise stated,
$\lambda_f=1.0$, $\lambda_r=0.5$, $\lambda_a=10^{-5}$, $m=0.1$, temperature
$\tau=1$, weight decay $0.01$, and gradient clipping at $1.0$.

\section{Experiments}

\subsection{Experimental setup}
\label{sec:experimental-setup}

\paragraph{Datasets.}
We evaluate on HotpotQA~\citep{yang2018hotpotqa},
2WikiMultihopQA~\citep{ho2020constructing}, and
MuSiQue~\citep{trivedi2022musique}, using the $1{,}000$-question evaluation sets
and corpora adopted by HippoRAG~\citep{hipporag,hipporag2}. Each question is
evaluated once, and we construct one entity--document graph per corpus. The
lead-time analysis uses the same evaluation questions solely for observation;
it is not used for parameter updates, threshold selection, or model selection.
Retriever training uses only HotpotQA and 2WikiMultihopQA training traces.
MuSiQue is therefore parameter-held-out for retriever optimization and model
selection.

\paragraph{Baselines.}
We compare RAPTOR~\citep{sarthi2024raptor},
GraphRAG~\citep{edge2024local}, LightRAG~\citep{guo2024lightrag},
HippoRAG~2~\citep{hipporag2}, Youtu-GraphRAG~\citep{dong2025youtu}, and
GFM-RAG~\citep{luo2025gfmrag}. Each baseline uses the same IRCoT-style
controller: a round retrieves from the question and prior thoughts, then
generates one additional thought. The controller stops after at most five rounds
or when the generator emits \texttt{\#\#\#}. Autoregressive variants use
Qwen2.5-7B. Dream-7B and Qwen2.5-7B receive matched supervised fine-tuning to
produce the same structured reasoning-trace interface (Appendix~\ref{app:sft}).
We also evaluate pretrained GFM-RAG with Dream-7B under this multi-round
controller. This DLM IRCoT baseline has neither entity-event triggering nor
event-conditioned retriever training. LADDER instead refreshes graph evidence
within a single denoising trajectory when a new entity emerges.

\paragraph{Metrics, and implementation.}
Table~\ref{tab:main} compares end-to-end systems: autoregressive baselines and
GFM-RAG (IRCoT) use thought-level multi-round retrieval, whereas LADDER
replaces this sequential loop with event-driven refreshes during parallel
denoising. Tables~\ref{tab:trigger} and~\ref{tab:trigger-full} instead hold a
Dream-7B trajectory and a fully event-conditioned GFM retriever fixed, changing
only the refresh policy. We report EM, F1, question-averaged Recall@$k$,
per-question wall-clock time, and denoising steps to convergence. Latency is
measured on a single NVIDIA H20 GPU with a warmed graph index. Dream-7B uses at
most $100$ denoising steps and $\tau_c=0.85$; GFM-RAG-8M uses
all-mpnet-base-v2 entity linking and $K=7$. Matching uses closed-set $n$-grams
of length at most $6$ with cumulative seeds.
\begin{table}[t]
\caption{End-to-end retrieval recall (\%). Autoregressive baselines and
GFM-RAG (IRCoT) use multi-round thought-level retrieval; LADDER uses
event-driven refreshes within one denoising trajectory. \textbf{Bold} marks the
best value in each column.}
\label{tab:recall}
\vspace{3pt}
\centering
\small
\resizebox{\textwidth}{!}{%
\setlength{\tabcolsep}{5pt}%
\begin{tabular}{l rrr rrr rrr}
\toprule
& \multicolumn{3}{c}{HotpotQA} & \multicolumn{3}{c}{2WikiMultihopQA} & \multicolumn{3}{c}{MuSiQue} \\
\cmidrule(lr){2-4}\cmidrule(lr){5-7}\cmidrule(l){8-10}
Method & R@2 & R@5 & R@7 & R@2 & R@5 & R@7 & R@2 & R@5 & R@7 \\
\midrule
\multicolumn{10}{l}{\footnotesize\emph{Autoregressive agent (Qwen2.5-7B)}}\\
RAPTOR~\citep{sarthi2024raptor} & 30.1 & 56.8 & 65.2 & 43.8 & 66.1 & 73.5 & 17.6 & 31.9 & 39.1 \\
LightRAG~\citep{guo2024lightrag} & 28.4 & 54.1 & 62.8 & 41.6 & 63.4 & 71.1 & 16.1 & 29.8 & 36.9 \\
GraphRAG~\citep{edge2024local} & 26.7 & 51.2 & 60.1 & 39.5 & 60.6 & 68.4 & 14.9 & 27.6 & 34.7 \\
HippoRAG~2~\citep{hipporag2} & 32.6 & 59.6 & 68.1 & 46.7 & 70.8 & 77.9 & 21.6 & 37.6 & 41.8 \\
Youtu-GraphRAG~\citep{dong2025youtu} & 33.4 & 61.1 & 69.6 & \textbf{47.8} & 72.1 & 79.4 & \textbf{22.3} & \textbf{38.4} & \textbf{42.7} \\
GFM-RAG~\citep{luo2025gfmrag} & 31.6 & 58.2 & 66.7 & 45.5 & 69.2 & 76.5 & 18.7 & 33.2 & 40.3 \\
\midrule
\multicolumn{10}{l}{\footnotesize\emph{Diffusion agent (Dream-7B)}}\\
GFM-RAG (IRCoT) & 27.5 & 53.0 & 61.0 & 45.3 & 68.3 & 73.3 & 20.4 & 36.4 & 40.1 \\
\textbf{\textsc{Ladder}} & \textbf{35.0} & \textbf{64.4} & \textbf{72.9} & 47.5 & \textbf{80.1} & \textbf{86.9} & 21.1 & 37.2 & 41.4 \\
\bottomrule
\end{tabular}}
\end{table}

\begin{table}[t]
\caption{End-to-end multi-hop QA performance. Autoregressive baselines and
GFM-RAG (IRCoT) use multi-round thought-level retrieval; LADDER uses
event-driven refreshes within one denoising trajectory. EM and F1 are
percentages; time is per-question wall-clock seconds ($\downarrow$).
\textbf{Bold} marks the best value in each column.}
\label{tab:main}
\vspace{3pt}
\centering
\small
\resizebox{\textwidth}{!}{%
\setlength{\tabcolsep}{5pt}%
\begin{tabular}{l rrr rrr rrr rrr}
\toprule
& \multicolumn{3}{c}{HotpotQA} & \multicolumn{3}{c}{2WikiMultihopQA} & \multicolumn{3}{c}{MuSiQue} & \multicolumn{3}{c}{Average} \\
\cmidrule(lr){2-4}\cmidrule(lr){5-7}\cmidrule(lr){8-10}\cmidrule(l){11-13}
Method & EM & F1 & Time$\downarrow$ & EM & F1 & Time$\downarrow$ & EM & F1 & Time$\downarrow$ & EM & F1 & Time$\downarrow$ \\
\midrule
\multicolumn{13}{l}{\footnotesize\emph{Autoregressive agent (Qwen2.5-7B)}}\\
RAPTOR & 41.0 & 46.9 & 6.35 & 58.3 & 63.5 & 6.72 & 15.3 & 27.0 & 7.18 & 38.2 & 45.8 & 6.75 \\
LightRAG & 36.5 & 43.1 & 7.46 & 54.2 & 59.8 & 7.83 & 13.8 & 23.8 & 8.51 & 34.8 & 42.2 & 7.93 \\
GraphRAG & 33.7 & 40.6 & 9.72 & 51.4 & 56.9 & 10.16 & 12.5 & 21.9 & 11.24 & 32.5 & 39.8 & 10.37 \\
HippoRAG~2 & 43.8 & 50.1 & 10.86 & 62.4 & 67.2 & 11.42 & 17.6 & 30.4 & 12.67 & 41.3 & 49.2 & 11.65 \\
Youtu-GraphRAG & 44.6 & 51.0 & 11.58 & 63.5 & 68.3 & 12.31 & 18.2 & \textbf{31.1} & 14.24 & 42.1 & 50.1 & 12.71 \\
GFM-RAG & 42.5 & 48.7 & 8.12 & 61.3 & 65.2 & 7.91 & 16.7 & 28.8 & 8.76 & 40.2 & 47.6 & 8.26 \\
\midrule
\multicolumn{13}{l}{\footnotesize\emph{Diffusion agent (Dream-7B)}}\\
GFM-RAG (IRCoT) & 41.3 & 47.5 & 11.54 & 60.8 & 64.7 & 10.54 & 16.6 & 29.9 & 11.83 & 39.6 & 47.4 & 11.30 \\
\textbf{\textsc{Ladder}} & \textbf{49.4} & \textbf{59.6} & \textbf{2.55} & \textbf{67.4} & \textbf{73.5} & \textbf{1.88} & \textbf{18.8} & 30.7 & \textbf{3.95} & \textbf{45.2} & \textbf{54.6} & \textbf{2.79} \\
\bottomrule
\end{tabular}}
\end{table}
\subsection{Main results}

Tables~\ref{tab:recall} and~\ref{tab:main} compare LADDER with six
autoregressive graph-retrieval pipelines and the Dream-7B GFM-RAG (IRCoT)
bridge baseline. LADDER achieves the best EM on all three datasets and the best
recall at nearly every cutoff. On HotpotQA and 2WikiMultihopQA, it improves over
the strongest baseline by $4.8$ and $3.9$ EM points, respectively. Averaged
over datasets, it improves EM from $39.6\%$ to $45.2\%$ and F1 from $47.4\%$ to
$54.6\%$ over GFM-RAG (IRCoT), while reducing latency from $11.30$ to
$2.79$\,s per question ($4.1\times$). It also exceeds the strongest
autoregressive pipeline, Youtu-GraphRAG, by $3.1$ average EM points while using
less than one quarter of its latency.

\begin{figure*}[t]
\centering
\includegraphics[width=0.94\textwidth]{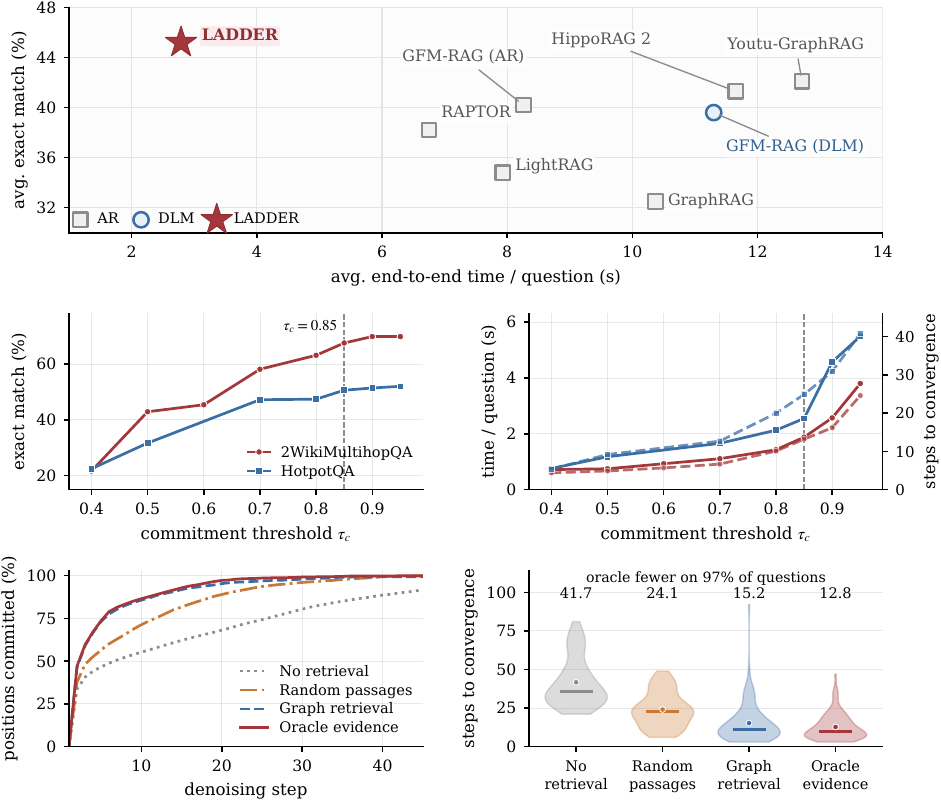}
\caption{Main results. \emph{Top:} average exact match versus average
per-question latency across the three datasets (Table~\ref{tab:main}), quality
increasing upward and latency leftward. \emph{Middle:} quality, latency, and
convergence steps as a function of the commitment threshold $\tau_c$ (the
vertical line marks $\tau_c=0.85$). \emph{Bottom:} convergence under four
evidence conditions on 2WikiMultihopQA, with (left) positions committed per
denoising step and (right) the distribution of steps to convergence.}
\label{fig:threshold}
\label{fig:pareto-main}
\label{fig:convergence}
\end{figure*}

Figure~\ref{fig:pareto-main} (top) summarizes this quality--latency trade-off.
LADDER occupies the upper-left region because it replaces thought-level
retrieval rounds with sparse event-level updates. MuSiQue is held out from
retriever optimization and model selection, so it evaluates parameter-held-out
transfer. LADDER attains the best MuSiQue EM ($18.8\%$ versus $18.2\%$ for
Youtu-GraphRAG) at $3.6\times$ lower latency, although it remains $0.4$ F1
points below the strongest baseline. This gap is consistent with the lower gold
coverage at events for longer reasoning chains (Figure~\ref{fig:lead}c).

\paragraph{Threshold sensitivity.}
Figure~\ref{fig:threshold} (middle) sweeps $\tau_c$ from $0.4$ to $0.95$ on
HotpotQA and 2WikiMultihopQA. EM plateaus above $0.8$, whereas latency and
convergence steps increase with the threshold. We therefore use one shared
operating point, $\tau_c=0.85$. Retrieval quality changes by less than two R@7
points over the sweep because entity events, rather than the number of denoising
steps, determine refreshes.

\subsection{Evidence quality accelerates denoising}
\label{sec:convergence}

A DLM commits masked positions in parallel once their confidence exceeds the
sampler threshold. Better evidence should resolve dependencies, commit more
positions per step, and shorten the trajectory. We test this prediction by
fixing the model, questions, and decoding hyperparameters while varying only the
$K$ prompt passages: no evidence, random corpus passages, graph retrieval, or
oracle evidence. Oracle prompts place gold documents first and pad the remaining
context with retrieved passages.

Figure~\ref{fig:convergence} (bottom) confirms a monotone effect: convergence
requires $41.7$ steps without evidence, $24.1$ with random passages, $15.2$
with graph retrieval, and $12.8$ with oracle evidence. Oracle evidence uses
fewer steps than no retrieval on $97\%$ of questions. Random context removes
$42\%$ of steps, and graph retrieval removes a further $37\%$ relative to random
context at fixed $K$. The same ordering holds on HotpotQA and MuSiQue, although
the gap narrows for longer chains (Appendix~\ref{app:convergence}). Thus, the
end-to-end latency gain reflects faster denoising convergence, not merely fewer
retrieval calls.

\subsection{Ablations}
\label{sec:ablations}

\begin{table*}[t]
\centering
\begin{minipage}[t]{0.48\textwidth}
\centering
\captionof{table}{(a) Single-trajectory refresh schedules, averaged over three
datasets.}
\label{tab:trigger}
\vspace{3pt}
\small
\begin{tabular*}{\linewidth}{@{\extracolsep{\fill}}lrrr}
\toprule
Schedule & Time$\downarrow$ & R@7 & EM \\
\midrule
Retrieve-once    & \textbf{2.55} & 55.6 & 37.9 \\
Every $5$ steps  & 2.98 & 62.9 & 42.6 \\
Per-step         & 4.38 & 65.8 & 44.5 \\
Random ($15\%$)  & 2.81 & 59.7 & 40.4 \\
\midrule
\textbf{\textsc{Ladder}} & 2.79 & \textbf{67.1} & \textbf{45.2} \\
\bottomrule
\end{tabular*}
\end{minipage}\hfill
\begin{minipage}[t]{0.48\textwidth}
\centering
\captionof{table}{(b) Training objectives, averaged over three datasets.}
\label{tab:training-ablation}
\vspace{3pt}
\small
\renewcommand{\arraystretch}{1.20}
\begin{tabular*}{\linewidth}{@{\extracolsep{\fill}}lrrr}
\toprule
Objective & R@7 & R$_{\text{disc}}$ & EM \\
\midrule
$\mathcal{L}_{\mathrm{retrieve}}$ only & 62.0 & 39.7 & 41.9 \\
$+\,\mathcal{L}_{\mathrm{lookahead}}$ & 65.4 & 49.8 & 44.2 \\
Full $-\,\mathcal{L}_{\mathrm{anchor}}$ & 65.4 & 50.3 & 44.2 \\
\midrule
\textbf{Full (\textsc{Ladder})} & \textbf{67.1} & \textbf{52.1} & \textbf{45.2} \\
\bottomrule
\end{tabular*}
\end{minipage}
\end{table*}

Table~\ref{tab:trigger} fixes the generator, retriever, and evidence budget
while varying the refresh policy. At comparable latency, LADDER improves over
\textsc{Random} by $7.4$ R@7 and $4.8$ EM. Per-step retrieval is $57\%$ slower
and remains worse, showing that repeated retrieval without a changed entity set
adds little useful evidence. The same
ordering holds on MuSiQue (Table~\ref{tab:trigger-full}).

Table~\ref{tab:training-ablation} fixes the LADDER schedule and varies the
retriever objective. Discovery recall, R$_{\text{disc}}$, measures the fraction
of gold documents absent from memory that an event retrieves.
$\mathcal{L}_{\mathrm{lookahead}}$ adds $10.1$ discovery points by supervising
missing evidence. Adding $\mathcal{L}_{\mathrm{retain}}$ without
$\mathcal{L}_{\mathrm{anchor}}$ leaves average R@7 and EM unchanged and degrades
MuSiQue transfer (Table~\ref{tab:training-full}). The final $+1.7$ R@7 and
$+1.0$ EM arise after adding $\mathcal{L}_{\mathrm{anchor}}$, indicating that
anchoring stabilizes event-conditioned retention around the pretrained
transferable representation.

\section{Conclusion}

LADDER turns graph-linkable entities in an intermediate diffusion draft into
retrieval events before their token spans are committed. This replaces
thought-level retrieval rounds with sparse graph updates: retrieval fires on
$15\%$ of denoising steps, mean EM improves from $39.6\%$ to $45.2\%$ over the
Dream-7B GFM-RAG (IRCoT) baseline, and end-to-end latency decreases by
$4.1\times$. Event-conditioned training adapts graph propagation to incomplete,
revisable queries while anchoring it to the pretrained transferable
representation. Since retriever updates use only HotpotQA and 2WikiMultihopQA
traces, MuSiQue provides a parameter-held-out zero-shot transfer evaluation.

\section{Limitations}
The closed-set entity matcher is intentionally simple, favoring low latency
and determinism over coverage; it does not yet link paraphrases, pronouns, or
entities absent from the corpus graph. Our experiments are also confined to
English multi-hop QA with the Dream-7B backbone, which reflects benchmark
availability rather than a constraint of the method: the event-driven trigger
and graph-retrieval mechanism are orthogonal to the denoising sampler and the
source language, so we expect \textsc{Ladder} to extend readily to other DLM
backbones and non-English multi-hop QA, and leave this validation to future
work.

\section{Ethics Statement}
This work is basic research on retrieval-augmented language models using
publicly available multi-hop QA benchmarks; it involves no human subjects or
private data collection. A potential limitation is that learned retrieval and
generation components may inherit factual errors, biases, or robustness issues
from their underlying corpora and models.

\section{AI Use Statement}
We used generative AI tools to improve the clarity and polish of the writing and
to assist with retrieval and discovery of related literature. We did not use
generative AI for research ideation or execution, drafting sections of the
paper, generating synthetic data, or proving mathematical claims. All
AI-assisted suggestions and retrieved references were reviewed and verified by
the authors, who take full responsibility for the final manuscript.

\bibliography{iclr2027_conference}
\bibliographystyle{unsrtnat}

\appendix

\section{GNN forward and differentiable document scoring}
\label{app:gnn-forward}

\paragraph{Inputs and seed initialization.}
We retain the GFM-RAG NBFNet-style reasoning architecture and fine-tune its
reasoning parameters on event traces. For an event query $u_t$, the frozen text
encoder produces $h_t\in\mathbb{R}^{d_h}$. Learned projections map the query
and fixed relation attributes to the NBFNet hidden dimension $d$,
\begin{equation}
 q_t=\phi_q(h_t)\in\mathbb{R}^{d}, \qquad
 R=\phi_r(\mathrm{rel\_attr})\in\mathbb{R}^{|\mathcal R|\times d}.
\label{eq:gnn-input}
\end{equation}
Let $m(E_t)\in\{0,1\}^{|\mathcal V_E|}$ be the cumulative multi-hot seed mask
over entity nodes. The initial node state is the outer product
\begin{equation}
 X_t^{(0)}=m(E_t)\otimes q_t,
 \qquad X_{t,v}^{(0)}=m_v(E_t)q_t,
\label{eq:seed-initialization}
\end{equation}
so every confirmed seed receives the same event-query representation and all
non-seed entities receive the zero boundary state.

\paragraph{Relation-aware propagation.}
The query-dependent NBFNet reasoner applies $L$ Bellman--Ford-style graph layers
while retaining the original boundary state at every layer. Writing
$H_t^{(0)}=X_t^{(0)}$, the recurrence is
\begin{equation}
 H_t^{(\ell)}=\mathcal B_\ell\bigl(H_t^{(\ell-1)},X_t^{(0)},q_t,R,\mathcal G\bigr),
 \qquad \ell=1,\ldots,L,
\label{eq:nbf-recurrence}
\end{equation}
where $\mathcal B_\ell$ performs relation-conditioned message passing over the
entity graph, aggregates incoming messages, and combines them with the boundary
condition. Residual connections are applied when input and output dimensions
match. The readout concatenates the configured hidden state(s) with $q_t$ and
maps them to scalar entity scores,
\begin{equation}
 z_t=\rho_\theta\bigl(H_t^{(1:L)},q_t\bigr)\in\mathbb{R}^{|\mathcal V_E|}.
\label{eq:gnn-forward}
\end{equation}
This formulation makes the event mask affect propagation only through
$X_t^{(0)}$; the graph topology and relation attributes are fixed across events.

\paragraph{Document readout and optimization.}
The fixed entity--document incidence matrix maps $z_t$ to the dense document
scores in Equation~\ref{eq:gfm}. During training, every document receives a
differentiable score before the listwise and hinge losses are evaluated. Thus,
gradients from $\mathcal{L}_{\mathrm{retrieve}}$,
$\mathcal{L}_{\mathrm{lookahead}}$, and $\mathcal{L}_{\mathrm{retain}}$ flow
through document aggregation, entity readout, NBFNet layers, and the query and
relation projections. At inference, we restore GFM-RAG's pretrained hard Top-20
entity ranker before selecting the final top-$K$ documents. The graph structure,
entity--document incidence matrix, text encoder, scheduler, and entity matcher
remain frozen; only the GFM reasoning model and its projection/readout parameters
are optimized.

\section{Generator fine-tuning}
\label{app:sft}

LADDER requires intermediate drafts to expose interpretable entities, so we
fine-tune Dream-7B to produce structured multi-hop reasoning traces rather than
relying on prompting alone. Qwen2.5-7B receives identical supervision data, a
matching output format, and the same training configuration, so both
generators share a matched reasoning-trace interface; differences reported in
the main paper are therefore attributable to retrieval control rather than
generator capability. Each training example pairs a question and its
supporting documents with a step-by-step trace ending in
\texttt{\#\#\# [answer]}, matching the shared IRCoT-style output format.

\paragraph{Output-mode analysis.}
Following the base-versus-fine-tuned output audit of
SARDI~\citep{juenger2026sardi}, we categorize outputs on a 500-question
2WikiMultihopQA subset under static $K=7$ retrieval (the \textsc{Retrieve-once}
condition in Table~\ref{tab:trigger-full}) as correct with a structured trace,
correct without one, or wrong (including format failures such as a missing
\texttt{\#\#\#} marker).

Base Dream-7B rarely emits a structured trace under prompting alone: it is
correct with a trace on only $1\%$ of questions (with another $23\%$ correct
but trace-free) and wrong otherwise. Base Qwen2.5-7B reasons far more often
($29\%$ correct with a trace) but still falls well short of either fine-tuned
model. After fine-tuning, both generators reason in nearly every output
($58\%$ and $59\%$ correct with a trace for Dream-7B and Qwen2.5-7B,
respectively, with the remainder wrong and no trace-free correct answers) and
land within one point of each other, consistent with the $57.6\%$
static-retrieval EM in Table~\ref{tab:trigger-full}. This supports treating
fine-tuning as a matched interface rather than a source of asymmetric
capability between generators.

\paragraph{Fine-tuning configuration.}
We fine-tune Dream-7B and Qwen2.5-7B with an identical configuration: $3$ training
epochs, learning rate $2\times10^{-6}$, global batch size $256$, micro batch size
$16$ per GPU, gradient accumulation steps $2$, maximum sequence length $2{,}048$,
and the AdamW optimizer. Training is performed on eight NVIDIA H20 GPUs and
inference on a single NVIDIA H20 GPU.

\section{Entity lead time statistics}
\label{app:lead}

Table~\ref{tab:lead} reports the per-dataset statistics underlying the
lead-time claim in Section~\ref{sec:motivation} and Figure~\ref{fig:lead}.

\begin{table}[h]
\caption{Entity span-level lead time, in denoising steps, over $1{,}000$ test
questions per dataset. An entity is committed when all of its token positions
are independently finalised by the sampler. Reproduced across nine shards (three
per dataset), where the $\text{Lead}>0$ rate stays within $85.8\%$--$88.8\%$.}
\label{tab:lead}
\vspace{3pt}
\centering
\small
\begin{tabular}{lrrrrr}
\toprule
Dataset & \#Entities & Mean & Median & P90 & $\text{Lead}>0$ \\
\midrule
2WikiMultihopQA & 1{,}892 & 5.71 & 3.0 & 15 & \textbf{88.8\%} \\
HotpotQA        & 3{,}090 & 9.35 & 5.0 & 24 & \textbf{88.0\%} \\
MuSiQue         & 4{,}524 & 9.62 & 6.0 & 24 & \textbf{88.5\%} \\
\bottomrule
\end{tabular}
\end{table}

Median lead time is shorter than the mean on all three datasets, indicating a
right-skewed distribution: most entities surface only a few steps before
commitment, while a minority emerge much earlier, as reflected in the P90
column. The consistently high $\text{Lead}>0$ rate across shards indicates that
early emergence is a stable property of the sampler rather than an artifact of
a particular evaluation split.

\section{Trigger quality of the entity matcher}
\label{app:trigger-quality}

We report how often an entity event yields a useful retrieval on all three
benchmarks. A retrieval is a real GNN forward (a cache hit fires no retrieval
when no new entity appears); we call it \emph{useful} when it increases the
number of gold documents in the top-$K$ list. Since a retrieved gold document is
retained, we measure the rate over triggers fired before full gold coverage,
where an increase is still possible.

\begin{table}[h]
\caption{Trigger quality across benchmarks. \emph{Triggers/q} is the mean
number of real graph retrievals per question; \emph{Useful-trigger rate} is the
fraction, among triggers before full gold coverage, that increase the number of
gold documents in the top-$K$ list.}
\label{tab:trigger-quality}
\vspace{3pt}
\centering
\small
\begin{tabular}{lcc}
\toprule
Dataset & Triggers/q & Useful-trigger rate \\
\midrule
HotpotQA        & $3.17$ & $53.3\%$ \\
2WikiMultihopQA & $2.07$ & $79.0\%$ \\
MuSiQue         & $4.35$ & $42.6\%$ \\
\bottomrule
\end{tabular}
\end{table}

The matcher remains sparse, with $2.07$--$4.35$ retrievals per question. Its
useful-trigger rate is highest on 2WikiMultihopQA and lowest on MuSiQue,
consistent with longer reasoning chains spreading the remaining evidence across
more retrieval events.

\section{Per-dataset ablation results}
\label{app:ablations}

Tables~\ref{tab:trigger} and~\ref{tab:training-ablation} report averages for readability.
Table~\ref{tab:trigger-full} and Table~\ref{tab:training-full} give the
corresponding per-dataset numbers, from which those averages are computed.
Table~\ref{tab:seed-policy} additionally reports the seed-policy ablation
referenced in Section~\ref{sec:graph-retrieval}.

\begin{table}[h]
\caption{Cumulative versus new-only seeding of the GNN start mask, per dataset,
with the trigger and all other settings fixed. Cumulative (\textsc{Ladder}'s
default) seeds every entity confirmed so far, $E_t$; new-seed uses only the
entities that triggered the current event, $\Delta E_t$.}
\label{tab:seed-policy}
\vspace{3pt}
\centering
\small
\begin{tabular}{l rr rr rr}
\toprule
& \multicolumn{2}{c}{HotpotQA} & \multicolumn{2}{c}{2WikiMultihopQA} & \multicolumn{2}{c}{MuSiQue} \\
\cmidrule(lr){2-3}\cmidrule(lr){4-5}\cmidrule(l){6-7}
Seed policy & R@7 & EM & R@7 & EM & R@7 & EM \\
\midrule
New-seed only ($\Delta E_t$) & 69.1 & 46.8 & 82.7 & 63.9 & 39.2 & 17.1 \\
\textbf{Cumulative} ($E_t$)  & \textbf{72.9} & \textbf{49.4} & \textbf{86.9} & \textbf{67.4} & \textbf{41.4} & \textbf{18.8} \\
\bottomrule
\end{tabular}
\end{table}

Cumulative seeding outperforms new-seed propagation on every dataset, most
notably on 2WikiMultihopQA ($+4.2$ R@7, $+3.5$ EM), confirming that carrying
earlier confirmed entities forward as graph seeds provides persistent value
beyond the entities that triggered the current event.

\begin{table}[h]
\caption{Single-trajectory refresh schedules under the same Dream-7B backbone
and full event-conditioned GFM retriever. Time is per-question end-to-end
wall-clock seconds ($\downarrow$). These ablations isolate refresh policy and
use a different execution unit from the multi-round IRCoT baseline in
Table~\ref{tab:main}. \textbf{Bold} marks the best value in each column.}
\label{tab:trigger-full}
\vspace{3pt}
\centering
\small
\resizebox{\textwidth}{!}{%
\setlength{\tabcolsep}{4pt}%
\begin{tabular}{l rrrr rrrr rrrr}
\toprule
& \multicolumn{4}{c}{HotpotQA} & \multicolumn{4}{c}{2WikiMultihopQA} & \multicolumn{4}{c}{MuSiQue} \\
\cmidrule(lr){2-5}\cmidrule(lr){6-9}\cmidrule(l){10-13}
Schedule & Time$\downarrow$ & R@7 & EM & F1 & Time$\downarrow$ & R@7 & EM & F1 & Time$\downarrow$ & R@7 & EM & F1 \\
\midrule
Retrieve-once ($t=0$)      & \textbf{2.28} & 61.9 & 42.1 & 51.8 & \textbf{1.75} & 71.2 & 57.6 & 62.4 & \textbf{3.62} & 33.6 & 13.9 & 24.1 \\
Fixed, every $10$ steps    & 2.41 & 66.8 & 45.4 & 55.1 & 1.85 & 76.8 & 61.2 & 66.3 & 3.80 & 36.2 & 15.4 & 26.3 \\
Fixed, every $5$ steps     & 2.67 & 69.7 & 47.2 & 57.0 & 2.05 & 80.5 & 63.8 & 69.1 & 4.21 & 38.5 & 16.8 & 28.2 \\
Per-step                   & 3.94 & 72.4 & 48.8 & 58.9 & 3.02 & 84.1 & 66.2 & 72.0 & 6.18 & 41.0 & 18.4 & 30.3 \\
\midrule
Random, firing rate $15\%$ & 2.52 & 66.1 & 44.8 & 54.5 & 1.94 & 76.1 & 60.6 & 65.7 & 3.97 & 36.8 & 15.9 & 26.9 \\
Conf.-drop~\citep{jiang2023active} & 2.49 & 68.9 & 46.5 & 56.2 & 1.91 & 81.7 & 64.9 & 70.2 & 3.88 & 38.9 & 17.2 & 28.6 \\
\midrule
\textbf{\textsc{Ladder}} & 2.55 & \textbf{72.9} & \textbf{49.4} & \textbf{59.6} & 1.88 & \textbf{86.9} & \textbf{67.4} & \textbf{73.5} & 3.95 & \textbf{41.4} & \textbf{18.8} & \textbf{30.7} \\
\bottomrule
\end{tabular}}
\end{table}

\begin{table}[h]
\caption{Training objectives, per dataset. \textbf{Bold} marks the best value
in each column.}
\label{tab:training-full}
\vspace{3pt}
\centering
\small
\resizebox{\textwidth}{!}{%
\setlength{\tabcolsep}{4pt}%
\begin{tabular}{ll rrrr rrrr rrrr}
\toprule
& & \multicolumn{4}{c}{HotpotQA} & \multicolumn{4}{c}{2WikiMultihopQA} & \multicolumn{4}{c}{MuSiQue} \\
\cmidrule(lr){3-6}\cmidrule(lr){7-10}\cmidrule(l){11-14}
& Objective & R@7 & R$_{\text{disc}}$ & EM & F1 & R@7 & R$_{\text{disc}}$ & EM & F1 & R@7 & R$_{\text{disc}}$ & EM & F1 \\
\midrule
(1) & $\mathcal{L}_{\mathrm{retrieve}}$ only & 66.4 & 42.7 & 45.2 & 54.0 & 78.9 & 50.2 & 63.4 & 68.5 & 40.6 & 26.1 & 17.2 & 30.0 \\
(2) & $+\,\mathcal{L}_{\mathrm{lookahead}}$ & 70.8 & 53.1 & 48.1 & 57.8 & 84.2 & 62.8 & 66.1 & 71.8 & 41.1 & 33.4 & 18.3 & 30.4 \\
(3) & Full (\textsc{Ladder}) & \textbf{72.9} & \textbf{55.8} & \textbf{49.4} & \textbf{59.6} & \textbf{86.9} & \textbf{65.7} & \textbf{67.4} & \textbf{73.5} & \textbf{41.4} & \textbf{34.9} & \textbf{18.8} & \textbf{30.7} \\
\midrule
(4) & Full $-\,\mathcal{L}_{\mathrm{anchor}}$ & 71.2 & 54.2 & 48.3 & 58.1 & 85.1 & 64.0 & 66.3 & 72.1 & 39.8 & 32.7 & 17.9 & 29.6 \\
\bottomrule
\end{tabular}}
\end{table}

\section{Retrieval-budget sensitivity}
\label{app:k-sensitivity}

Section~\ref{sec:graph-retrieval} fixes the retrieval budget at $K=7$ for all
main-text experiments. Table~\ref{tab:k-sensitivity} sweeps $K\in\{5,7,10,15\}$
on 2WikiMultihopQA and MuSiQue, holding the generator, retriever checkpoint,
and event-triggering policy fixed and varying only the number of top-ranked
documents placed in the prompt at each retrieval event.

\begin{table}[h]
\caption{Sensitivity of \textsc{Ladder} to the retrieval budget $K$ on
2WikiMultihopQA and MuSiQue, with the generator, retriever, and trigger policy
held fixed. EM and F1 are percentages; R@$K$ is the final-state recall at the
evaluated budget; \emph{Triggers/q} is the mean number of real graph
retrievals per question; latency is per-question wall-clock seconds
($\downarrow$).}
\label{tab:k-sensitivity}
\vspace{3pt}
\centering
\small
\begin{tabular}{lrrrrrr}
\toprule
Dataset & $K$ & EM & F1 & R@$K$ (final) & Triggers/q & Time$\downarrow$ \\
\midrule
\multirow{4}{*}{2WikiMultihopQA}
& $5$        & 59.0 & 64.6 & 80.1 & 2.07 & 1.56 \\
& $7$        & 67.4 & 73.5 & 86.9 & 2.07 & 1.88 \\
& $10$       & 73.0 & 76.3 & 92.1 & 2.08 & 2.70 \\
& $15$       & 71.0 & 75.7 & 93.9 & 2.09 & 3.55 \\
\midrule
\multirow{4}{*}{MuSiQue}
& $5$        & 15.0 & 25.1 & 37.2 & 4.14 & 3.17 \\
& $7$        & 18.8 & 30.7 & 41.4 & 4.35 & 3.95 \\
& $10$       & 19.0 & 28.8 & 41.5 & 4.39 & 5.80 \\
& $15$       & 23.0 & 32.0 & 47.2 & 4.53 & 7.55 \\
\bottomrule
\end{tabular}
\end{table}

Recall improves monotonically with $K$ on both datasets, as expected: a larger
budget can only add documents to the retrieved set. EM and F1 do not track
recall monotonically, however. On 2WikiMultihopQA, $K=10$ attains the highest
EM and F1 in the sweep, slightly above the $K=7$ operating point, while
$K=15$ EM falls back despite higher recall, indicating that additional
lower-ranked passages can dilute the prompt with distracting context. On
MuSiQue, EM and F1 increase monotonically with $K$ up to $15$, consistent with
its longer reasoning chains and lower per-event gold coverage
(Section~\ref{sec:motivation}), which leave more headroom for extra evidence
to help. In both cases, however, latency grows steeply with $K$ (by
$1.9\times$ from $K=7$ to $K=15$ on 2WikiMultihopQA and MuSiQue alike), while
the trigger count \emph{Triggers/q} stays nearly flat, confirming that the
extra cost comes from scoring and prompting more documents per event rather
than from additional retrieval events. We therefore use $K=7$ as a shared
operating point that is within one to two EM points of the best value on each
dataset while avoiding the latency growth of larger budgets.

\section{Convergence on the remaining datasets}
\label{app:convergence}

\begin{figure}[h]
\centering
\includegraphics[width=\textwidth]{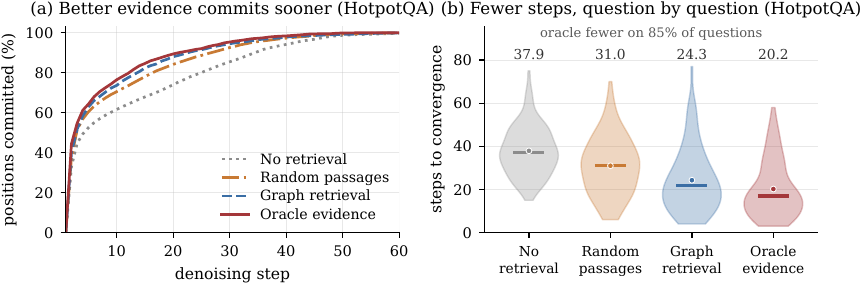}
\caption{Convergence on HotpotQA, in the format of
Figure~\ref{fig:convergence}.}
\label{fig:convergence-hotpot}
\end{figure}

Section~\ref{sec:convergence} reports the evidence-quality experiment on
2WikiMultihopQA. Table~\ref{tab:convergence-all} repeats it on HotpotQA and
MuSiQue under identical conditions: same checkpoint, same question set per
dataset, same decoding hyper-parameters, and only the $K=7$ passages in the
prompt varying. Figure~\ref{fig:convergence-hotpot} plots the HotpotQA curves in
the same format as the main text.

\begin{table}[h]
\caption{Steps to convergence by evidence condition. \emph{Paired} is the
share of questions on which oracle evidence converges in strictly fewer steps
than no retrieval.}
\label{tab:convergence-all}
\vspace{3pt}
\centering
\small
\begin{tabular}{llrrrr}
\toprule
Dataset & Evidence & Mean & SD & Median & Paired \\
\midrule
\multirow{4}{*}{2WikiMultihopQA}
 & No retrieval      & 41.70 & 15.68 & 36.0 & --- \\
 & Random passages   & 24.05 & 11.30 & 23.0 & --- \\
 & Graph retrieval   & 15.24 & 12.57 & 11.0 & --- \\
 & Oracle evidence   & \textbf{12.76} & 8.43 & 10.0 & $97\%$ \\
\midrule
\multirow{4}{*}{HotpotQA}
 & No retrieval      & 37.91 & 11.37 & 37.0 & --- \\
 & Random passages   & 31.00 & 13.54 & 31.0 & --- \\
 & Graph retrieval   & 24.30 & 14.73 & 22.0 & --- \\
 & Oracle evidence   & \textbf{20.21} & 12.99 & 17.0 & $85\%$ \\
\midrule
\multirow{4}{*}{MuSiQue}
 & No retrieval      & 42.39 & 11.08 & 41.0 & --- \\
 & Random passages   & 37.36 & 13.81 & 36.0 & --- \\
 & Graph retrieval   & 34.05 & 17.21 & 32.5 & --- \\
 & Oracle evidence   & \textbf{33.22} & 18.51 & 32.0 & $74\%$ \\
\bottomrule
\end{tabular}
\end{table}

The ordering of the four conditions is the same on all three datasets, but the
size of the effect is not, and it tracks the length of the reasoning chain.
Oracle evidence cuts the trajectory to $31\%$ of its unretrieved length on
2WikiMultihopQA, to $53\%$ on HotpotQA, and only to $78\%$ on MuSiQue, where the
gap between retrieved and oracle evidence also narrows to $0.8$ steps. This is
consistent with Figure~\ref{fig:lead}c: MuSiQue questions require the longest
chains and have the lowest fraction of trigger events at which the gold
documents are already available, so a single injection of evidence resolves a
smaller share of the draft's dependencies and more of the decoding work is
irreducible. The mechanism is therefore strongest exactly where multi-hop
evidence is most self-contained, and weakest where dependencies are deepest.

\section{Prompt templates}
\label{app:prompt}

\paragraph{RAG input format.}
All models receive input in the following format:
\begin{quote}
\ttfamily
Use ONLY the provided facts to answer the question.\\
Think step-by-step, then provide the final answer after the "\#\#\#" marker.\\[2pt]
Question:\\
\{question\}\\[2pt]
Facts:\\
\{facts\}
\end{quote}
where \texttt{\{question\}} is the input question and \texttt{\{facts\}}
contains the concatenated retrieved passages.

\paragraph{Expected output format.}
Models are trained to produce outputs in the following format:
\begin{quote}
\ttfamily
Step 1: [First reasoning step grounded in documents]\\
Step 2: [Second reasoning step]\\
...\\
\#\#\# [final answer]
\end{quote}
This format serves two purposes: it enables extraction of intermediate entities
from partial reasoning traces for self-augmenting retrieval, and it provides a
consistent extraction point (\#\#\#) for evaluation.

\section{Case study}
\label{app:case}

\begin{table}[h]
\caption{Case study: \emph{``What larger educational institution is the
university where Charlotte Brewer went, a part of?''} Gold documents:
\textsc{Charlotte Brewer} and \textsc{Hertford College, Oxford}. Bold tokens
in the denoising state are independently committed positions; the remainder is
the proxy's argmax fill of still-masked positions. GNN marks a forward pass;
\checkmark\ marks a scheduled retrieval call served by cached graph scores; and
$\varnothing$ marks a step where the unchanged query skips the retrieval call.
Gold is the number of gold documents in the retrieved top-$K$ out of the total.}
\label{tab:case}
\vspace{3pt}
\centering
\small
\begin{tabular}{@{}cccp{0.43\textwidth}c@{}}
\toprule
Step & Event & Retrieval & Denoising state $u_t$ & Gold \\
\midrule
$0$ & --- & fwd & \textit{(question only)} & $1/2$ \\
$1$ & \{college, Oxford, University\} & fwd & \textbf{Charlotte Brewer} is at H Soulsford College, Oxford. (2): \dots & $1/2$ \\
$2$ & --- & \checkmark & \textbf{Charlotte Brewer} is at professor offord College, Oxford. (2): \dots & $1/2$ \\
$3$ & --- & \checkmark & \textit{(wording drifts further; entity set unchanged)} & $1/2$ \\
$4$--$5$ & --- & $\varnothing$ & \textit{(proxy text identical to step $3$)} & $1/2$ \\
$6$ & --- & \checkmark & \textbf{Charlotte Brewer} is at professor offord College, Oxford. (2): \dots \textit{(still wrong)} & $1/2$ \\
$7$ & \{Hertford College\} & fwd & \textbf{Charlotte Brewer} is at Hertford College, Oxford. (2): \dots ert College Oxford \dots & $\mathbf{2/2}$ \\
$8$ & --- & \checkmark & \textbf{Charlotte Brewer is} College. (2): Hert College a constituent college of the University \#\#\# Oxford & $2/2$ \\
$9$ & --- & \checkmark & \textbf{Charlotte Brewer is at Hertford College, Oxford. (2): College a constituent college of the University \#\#\#} & $2/2$ \\
$10$ & --- & $\varnothing$ & \textbf{Charlotte Brewer is at Hertford College, Oxford. (2): College a constituent college of the University \#\#\# Oxford} & $2/2$ \\
\bottomrule
\end{tabular}
\end{table}
Table~\ref{tab:case} traces every denoising step of a single question from a
motivation-collection trajectory (Section~\ref{sec:motivation}), to make the
entity-emergence signal, the caching it enables, and its effect on retrieved
evidence concrete rather than aggregate. The question, \emph{``What larger
educational institution is the university where Charlotte Brewer went, a part
of?''}, is a bridge question: the second hop, which college Charlotte Brewer
attended, is absent from the query itself, so the two gold documents
(\textsc{Charlotte Brewer} and \textsc{Hertford College, Oxford}) cannot both be
retrieved from the question alone. At step $0$ the initial query retrieves one
of the two.

At step $1$ the proxy has drifted
to ``\dots is at H Soulsford College, Oxford\dots'', a plausible-sounding but
fabricated college name. The entity matcher extracts \emph{college}, \emph{Oxford}
and \emph{University} from this text. These are valid but generic graph nodes,
so they form a genuine event ($|\Delta E_t|\geq 1$) and refresh retrieval, yet
their diffuse propagation does not materially change the retrieved set and gold
coverage remains $1/2$. The fabricated string ``H Soulsford College'' itself is
not matched: the closed-vocabulary matcher (Section~\ref{sec:trigger}) admits
only canonical graph nodes, preventing nonexistent entities from entering the
seed mask. Between steps $2$ and $6$ the proxy's
second half continues to oscillate, reading ``professor offord College''
at step $2$, then stable but wrong through step $6$. During this span, no new
entity is matched, so the seed mask never changes and no GNN forward pass is
issued; two levels of caching absorb the five idle steps. When the proxy text
itself is byte-identical to the previous step (steps $4$ and $5$), the query
cache returns the previous passages immediately, at negligible cost. When the
text has drifted but the entity set has not (steps $2$, $3$ and $6$), the
closed-set matcher returns the cached graph-score vector; neither query encoding
nor document re-scoring is performed. These cache-served calls take
$0.0006$--$0.003$\,s, versus $0.055$\,s for the step-$1$ event that changes the
seed mask. Across the trajectory only $3$ of the $11$ steps issue a GNN forward
pass, while the remaining $8$ are served from cache or skipped outright, as
anticipated in Section~\ref{sec:graph-retrieval}.

At step $7$ the masked positions resolve to
``\dots is at Hertford College, Oxford\dots'', and the matcher extracts
\textsc{Hertford College}, the exact title of the missing gold document. The
entity is legible in the proxy two full steps before all of its token positions
are committed, illustrating the span-level lead-time distribution in
Figure~\ref{fig:lead}a. Because the entity is a real graph node, this event does
change the retrieved set: the GNN performs a forward pass (its third and last in
this trajectory,
after the unconditional retrieval at step $0$ and the false-start event at step
$1$) seeded with the accumulated entities
\{\emph{college}, \emph{Oxford}, \emph{University}, \textsc{Charlotte Brewer},
\textsc{Hertford College}\}, and \textsc{Hertford College, Oxford} enters the
retrieved set at rank $2$. Gold coverage reaches $2/2$ at this step and remains
there for the rest of denoising. All token positions of ``Hertford College''
are committed at step $9$, and the trajectory terminates at step $10$ with the
correct final answer, \emph{University of Oxford}.

This sparsity (only $3$ of $11$ steps issuing a GNN forward pass) is the
mechanism behind the $15\%$ average firing rate reported throughout the paper.
The one useful event, at step $7$, is also the one that could not have been
triggered by monitoring only committed positions, since the span ``Hertford
College'' is not fully committed until two steps later; a scheme that waits
for committed text would miss this window entirely, and a scheme that fires on
any token-level change in the proxy would instead fire on every one of steps
$2$, $3$ and $6$, where the
draft's wording shifts but no new entity, and hence no new evidence, is available.

\end{document}